\documentclass[conference]{IEEEtran}

\usepackage[table]{xcolor}
\usepackage{booktabs,calc,diagbox}
\usepackage{amsmath}
\usepackage[numbers,sort&compress]{natbib}
\usepackage{tikz}
\usetikzlibrary{positioning, chains, shapes.geometric, fit, shapes, arrows.meta, calc, backgrounds}

\newcommand{\keywords}[1]{\par\noindent\textbf{Keywords:} #1}

\newcommand{\cellc}{\cellcolor{gray!20}}
\newcommand\copyrighttext{
  \footnotesize \textcopyright 2025 Advanced Micro Devices, Inc. All rights reserved. AMD, the AMD Arrow logo, and combinations thereof are trademarks of Advanced Micro Devices, Inc. Other product names used in this publication are for identification purposes only and may be trademarks of their respective companies.}
\newcommand\copyrightnotice{
\begin{tikzpicture}[remember picture,overlay]
\node[anchor=south west,xshift=50pt, yshift=20pt] at (current page.south west) 
  {
    \parbox{\dimexpr\columnwidth-\fboxsep-\fboxrule\relax}{\copyrighttext}
    };
\end{tikzpicture}
}

\title{LCS: An AI-based Low-Complexity Scaler for Power-Efficient Super-Resolution of Game Content}

\author{Simon Pochinda, Momen K. Tageldeen, Mark Thompson, Tony Rinaldi, Troy Giorshev, \\Keith Lee, Jie Zhou, Frederick Walls*\\
Advanced Micro Devices, Inc. \\
*Corresponding author Frederick.Walls@amd.com 
} 

\begin{document}

\maketitle 
\copyrightnotice


\begin{abstract}
The increasing complexity of content rendering in modern games has led to a problematic growth in the workload of the GPU. In this paper, we propose an AI-based low-complexity scaler (LCS) inspired by state-of-the-art efficient super-resolution (ESR) models which could offload the workload on the GPU to a low-power device such as a neural processing unit (NPU). The LCS is trained on GameIR image pairs natively rendered at low and high resolution. We utilize adversarial training to encourage reconstruction of perceptually important details, and apply reparameterization and quantization techniques to reduce model complexity and size. In our comparative analysis we evaluate the LCS alongside the publicly available AMD hardware-based Edge Adaptive Scaling Function (EASF) and AMD~FidelityFX\texttrademark~Super Resolution~1 (FSR1) on five different metrics, and find that the LCS achieves better perceptual quality, demonstrating the potential of ESR models for upscaling on resource-constrained devices.
\end{abstract}


\section{Introduction}\label{sec:introduction}
Consumers increasingly demand higher resolution content and higher frame rates. Meanwhile, game rendering, including realistic physics and compute intensive operations such as ray tracing, is becoming increasingly complex. This has led to an increase in the workload on the GPU, which is unsustainable on resource constrained devices. Solutions such as AMD~FidelityFX\texttrademark~Super Resolution (FSR, AMD), Deep Learning Super Sampling (DLSS, NVIDIA), and Xe Super Sampling (XeSS, Intel) circumvent these rendering constraints by rendering game content at lower resolutions followed by an upscaling operation to transform the image dimensions to the target resolution. However, the complexity of these algorithms demands GPU resources, which is also compute- and power-intensive. In this paper, we explore offloading ML-based upscaling using a small and more power-efficient, low-complexity model (i.e., to an NPU or other low-power device). 
Content upscaling with low-complexity models has been demonstrated with several deep learning-based single-image super-resolution (SISR) models \cite{shi_2016_rtsr,Kong_2022_rlfn,chen_2022_esr} and efficient super-resolution \citep[ESR; see][for a review of the basic design choices]{Zamfir_2023_towardsrtsr} has been the focus of a number of recent conference workshop challenges \citep{Li_2022_esrchallenge,Li_2023_esrchallenge, Ren_2024_esrchallenge, Ren_2025_esrchallenge}. Participating teams in these workshop challenges typically apply a combination of efficient design choices and optimization techniques to achieve high model performance, including network pruning \citep{LeCun_1989_pruning, Han_2015_pruning, li_2017_pruning, liu_2018_pruning}, reparameterization \citep{ding_2021_reparameterization, Du_2022_reparameterization}, network quantization \citep{Gholami_2021_quant, Colbert_2023_quant}, neural architecture search \citep{elsken_2019_nas, White_2023_nas}, and knowledge distillation \citep{hinton_2015_kd, Romero_2015_kd}. The models developed during these workshop challenges represent the state-of-the-art in ESR and are typically scored using metrics like the average inference runtime, floating point operations per second (FLOPS), and number of model parameters, while required to maintain a minimum peak signal-to-noise ratio (PSNR) score. 

ESR training data usually consists of high-resolution (HR) and low-resolution (LR) image pairs such as DIV2K \citep{Agustsson_2017_div2k, Timofte_2017_div2k}, Flickr2K \citep{Timofte_2017_div2k}, and LSDIR \citep{Li_2023_lsdir}. Typically, these image pairs are generated by downsampling the HR images using bicubic interpolation and, optionally, adding distortions like noise and blurring to the LR images. However, this corruption process does not accurately simulate LR images generated by a game engine, and hence the models trained on these datasets do not perform well in the upscaling task targeting real gaming content \citep{Zhou_2025_gameir}. In addition, the commonly used L1 and L2 training objectives used in ESR also have limitations. As these loss functions minimize the Manhattan and Euclidean distance by averaging all possible outputs, they inherently maximize the PSNR metric. While this is effective for meeting the criteria of the ESR challenges, it can lead to blurry, perceptually dissatisfying images \citep{Pathak_2016_blurryloss, Isola_2017_blurryloss}.

In summary, this work focuses on designing a suitable low-complexity ESR model, training the model on natively rendered LR and HR image pairs, and identifying a training objective that produces perceptually satisfying images. To address these challenges, we propose a low-complexity scaler (LCS) based on Residual Local Feature Network \citep[RLFN;][]{Kong_2022_rlfn} and DIPNet \citep{Yu_2023_dipnet} utilizing adversarial training \citep{goodfellow_2014_gan} inspired by ESRGAN \citep{Wang_2018_esrgan} to encourage reconstruction of perceptually important details. Our model is trained on the GameIR dataset \citep{Zhou_2025_gameir}, which contains true LR and HR image pairs natively rendered by a game engine at each resolution. Additionally, we exploit reparameterization to reduce the model complexity, and quantization to further reduce the model size. To evaluate the performance of our LCS we conduct a comparative analysis with publicly available upscaling methods such as FSR1 and the AMD hardware-based EASF.

This paper is organized as follows: 
Section \ref{sec:methodology} describes the architecture, training details, and efficiency optimizations of our LCS. 
Section \ref{sec:results} presents the results of our experiments, comparing the performance of our approach with traditional upscaling methods. 
Finally, Section \ref{sec:disc_conc} concludes the paper and discusses future work.

\section{Methodology} \label{sec:methodology}

\subsection{Dataset and metrics} \label{sec:dataset}
In this work, we use the GameIR dataset \citep{Zhou_2025_gameir}, which contains both static and dynamic images (i.e., with and without moving vehicles) from the CARLA driving simulator rendered with Unreal Engine 4 \citep{Dosovitskiy_2017_carla}. The dataset consists of images for SR (LR-HR image pairs) and novel view synthesis (segmentation and depth maps). This work uses the SR dataset, which has 19,200 LR-HR image pairs rendered at 720p and 1440p respectively; however, to reduce training time and allow for more epochs, we use a smaller subset consisting of 1434 image pairs (1194 for training, 240 for validation).

We evaluate the performance of the LCS on several metrics: The peak signal-to-noise ratio (PSNR), structural similarity index measure \citep[SSIM;][]{Wang_2004_ssim}, natural image quality evaluator \citep[NIQE;][]{Mittal_2013_niqe}, just-objectionable-difference \citep[JOD;][]{Mantiuk_2024_colorvideovdp}, and learned perceptual image patch similarity \citep[LPIPS;][]{Zhang_2018_lpips}. 

\subsection{Network architecture}
In Figure \ref{fig:network} below we show the architecture of the LCS generator and its constituent blocks alongside the discriminator network used for adversarial training. The proposed LCS is based on the DIPNet architecture \citep{Yu_2023_dipnet}. DIPNet itself is based on the RLFN \citep{Kong_2022_rlfn} which improves on the Residual Feature Distillation Network \citep[RFDN;][]{Liu_2020_rfdn}. The modifications introduced in DIPNet includes exchanging the RLFN residual local feature blocks (RLFBs) with reparameterization residual feature blocks (RRFBs). In this work we also adopt RRFBs, but we replace the DIPNet residual reparameterization blocks (RRBs) inside the RRFBs with the residual in residual reparameterization blocks (RRRBs) from \cite{Du_2022_reparameterization} followed by a ReLU activation. The RRRBs enable feature extraction in a higher-dimensional space, increasing the learning capacity of the model. During inference, the RRRBs are reparameterized to a single 3x3 convolution, which significantly reduces the number of parameters and speeds up inference. Similar to DIPNet and RLFN, each RRFB in our model also includes enhanced spatial attention (ESA) blocks which serves as a low-complexity spatial attention module \citep{Liu_2020_rfanesa}. Our trained LCS model uses four consecutive RRFBs with 38 feature channels and a feature expansion factor of two.

\subsection{Adversarial training}
\begin{figure*}
\centering
\input{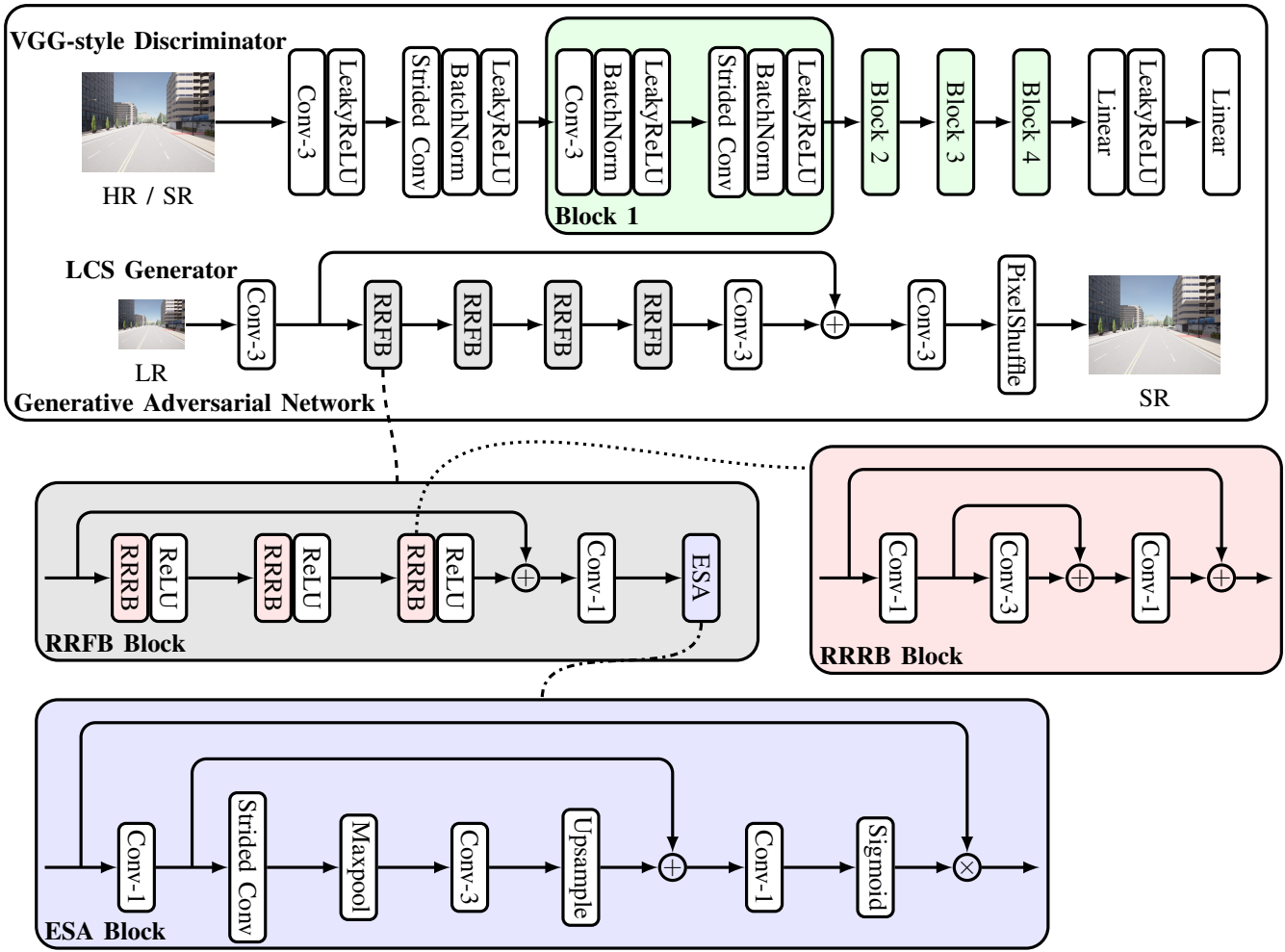}
\caption{Architecture of the low-complexity scaler and overview of reparameterization residual feature blocks (RRFB), residual in residual reparameterization blocks (RRRBs), and enhanced spatial attention blocks (ESA).}\label{fig:network}
\end{figure*}

Unlike DIPNet and RLFN, we train the LCS using adversarial training inspired by ESRGAN \citep{Wang_2018_esrgan}. Figure \ref{fig:network} shows the adversarial training setup. The setup is comprised of the LCS generator and the relativistic VGG-style discriminator \citep{jolicoeur_2018_rgan, Simonyan_2015_vgg, Wang_2018_esrgan}, which learns whether the SR image is more realistic than the HR image. We replace the ESRGAN generator with our LCS, and adopt the generator and discriminator loss functions from ESRGAN \citep{Wang_2018_esrgan}. The generator training objective combines adversarial loss with weight $\lambda=5\times 10^{-3}$, L1 loss with weight $\eta = 1\times 10^{-2}$, and a perceptual loss \citep{johnson_2016_percep}. Training is initialized with separate Adam optimizers \citep{Kingma_2015_adam} for the generator and discriminator, both with $\beta_1 = 0.9$ and $\beta_2 = 0.999$ and a learning rate of $1\times 10^{-4}$ that is halved at iteration [50k, 100k, 200k, 300k]. During training we augment the data with random horizontal flips, random rotations, and random crops of size $128\times 128$. The training runs for a total of 500k iterations with a batch size of 4, corresponding to $\sim 1700$ epochs on a single AMD~Instinct\texttrademark~MI210 GPU.

\subsection{Reparameterization and quantization}
To reduce the memory footprint and improve the inference speed of the LCS, we apply several optimizations. First, we use reparameterization \citep{ding_2021_reparameterization, Du_2022_reparameterization} which allows us to reduce the number of parameters in the RRFBs during inference. Specifically, this reduces the number of parameters in the LCS generator from $\sim 0.74$M to $\sim 0.21$M, while maintaining similar performance to the original model.

To further optimize the LCS we quantize the model. We adapt the LCS architecture with the equivalent quantization-aware layers in the quantization library Brevitas \citep{giuseppe_2025_brevitas} to quantize the model weights to 8-bit integers (INT8). This is done by first training the model in full 32-bit floating point (FP32) precision and then fine-tuning the model using quantization-aware training (QAT) \citep{Gholami_2021_quant, Colbert_2023_quant}. QAT improves the inference speed and, more importantly, significantly reduces the memory footprint of the model. A small memory footprint is crucial to constraining the silicon area.

\section{Results} \label{sec:results}
In this section we evaluate the model performance of our LCS, the EASF, and FSR1 scaling algorithms on the GameIR validation dataset \citep{Zhou_2025_gameir} as described in Section~\ref{sec:dataset}. Our comparative analysis includes common metrics such as PSNR and SSIM \citep{Wang_2004_ssim}, as well as perceptual metrics such as NIQE \citep{Mittal_2013_niqe}, JOD \citep{Mantiuk_2024_colorvideovdp}, and LPIPS \citep{Zhang_2018_lpips}.

\subsection{Qualitative evaluation}
Figure \ref{fig:comparison} below shows a side-by-side comparison of an upscaled validation image using the LCS, reparameterized LCS, reparameterized and quantized LCS, FSR1, and EASF. Subtle differences can be identified in the zoomed upscaled images:

\begin{figure*}
  \centering
  \includegraphics[width=\textwidth]{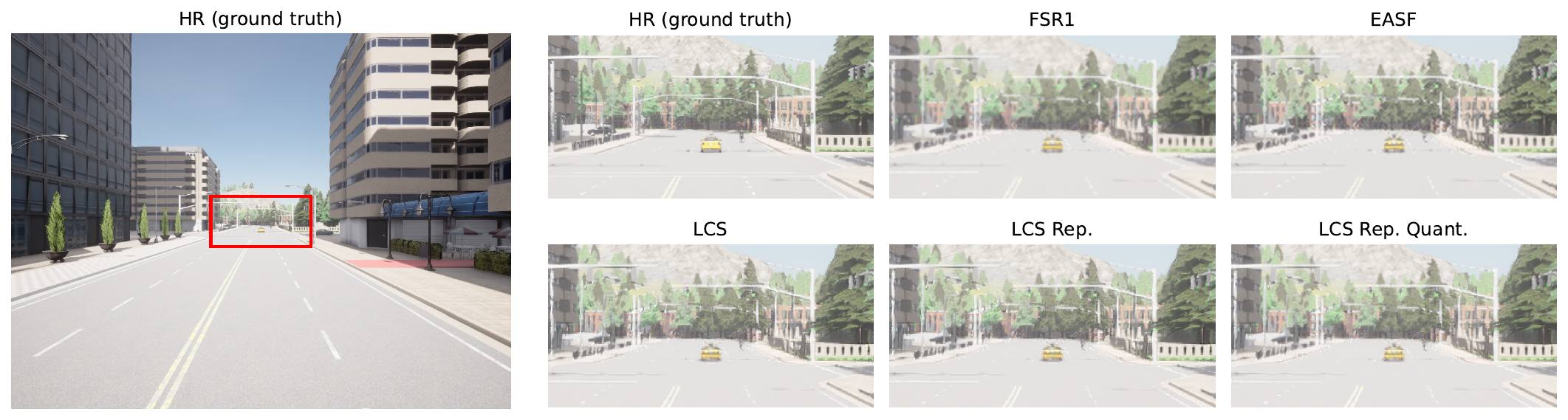}
  \caption{Comparison of the LCS, reparameterized LCS, reparameterized and quantized LCS, FSR1, and EASF on a GameIR dynamic validation image. 
  }\label{fig:comparison}
\end{figure*}

\begin{itemize}
  \item FSR1 appears to predict the mean of the pixel distribution well but produces a noticeably blurry output. It seems to struggle with particular areas such as the trees, while it reconstructs certain details, such as the back of the yellow car, relatively well. 
  \item EASF appears to reconstruct the pixel variance relatively well, realistically replicating the in-game noise grain better than FSR1.
  \item The LCS (FP32) model achieves a balance between smoothness and sharpness, reconstructing a satisfying level of noise grain while preserving the smoothness characteristic of the HR image.
  \item The reparameterized LCS (FP32) model is indistinguishable from the LCS (FP32) model, which is expected as the weights are combined at higher precision (FP64) to ensure consistent outputs.
  \item The reparameterized and quantized LCS (INT8) model shows comparable performance to the LCS and reparameterized LCS, suggesting that our quantization approach does not negatively impact the model performance.
\end{itemize}

Note that due to the nature of game rendering, the LR and HR image pairs in the GameIR dataset differ noticeably in the fine details. For example, the corresponding LR image (not shown) to the HR image in Figure~\ref{fig:comparison} includes a dark line below the taillights of the yellow car. Additionally, many HR and LR images exhibit slight color differences in some areas. Figure~\ref{fig:appendix_comparison} in Appendix~\ref{appendix:figs} highlights the differences in a side-by-side comparison including both the HR and LR image.

\subsection{Quantitative analysis}
Table \ref{tab:metrics} shows metric scores of the LCS, reparameterized LCS, reparameterized and quantized LCS, FSR1, and EASF alongside number of parameters, runtime, and multiply-accumulate operations (GMACs). Runtime was measured for LCS and reparameterized LCS on an AMD~Instinct\texttrademark~MI210 GPU as they were the models implemented in PyTorch with GPU compatible layers.

\begin{table}[h]
\centering
\renewcommand{\arraystretch}{1.3}
\caption{Comparative analysis of metrics alongside 68\% confidence intervals for the LCS, reparameterized LCS, reparameterized and quantized LCS, FSR1, and EASF. 
The best point estimates are highlighted in gray. A LR image of size $960\times 720$ was used to measure runtime and the total number of GMACs.
}\label{tab:metrics}
\tabcolsep=0.09cm
\begin{tabular}{@{} p{1cm} p{1.4cm} p{1.4cm} p{1.4cm} p{1.4cm} p{1.4cm} @{}}
  \toprule
  Metric                      &       LCS                       &       LCS Rep.                  &  LCS Rep. Quant.           &       FSR1&       EASF \\
  \midrule
  PSNR$\uparrow$ [dB]        &       $31.78^{+1.32}_{-1.17}$    &       $31.78^{+1.32}_{-1.17}$   &  $32.03^{+1.27}_{-1.18}$   & \cellc$32.67^{+1.20}_{-1.20}$   &       $32.38^{+1.28}_{-1.10}$   \\
  SSIM$\uparrow$             &       $0.908^{+0.023}_{-0.023}$  &       $0.908^{+0.023}_{-0.023}$ &  $0.911^{+0.022}_{-0.023}$ & \cellc$0.923^{+0.018}_{-0.019}$ &       $0.921^{+0.018}_{-0.019}$ \\
  NIQE$\downarrow$           & \cellc$3.43^{+0.37}_{-0.44}$     & \cellc$3.43^{+0.37}_{-0.44}$    &\cellc$3.43^{+0.38}_{-0.40}$&       $5.55^{+0.57}_{-0.55}$    &       $5.20^{+0.31}_{-0.31}$    \\
  JOD$\uparrow$              &       $8.50^{+0.12}_{-0.11}$     &       $8.50^{+0.12}_{-0.11}$    &\cellc$8.51^{+0.12}_{-0.11}$& \cellc$8.51^{+0.14}_{-0.13}$    &       $8.50^{+0.13}_{-0.10}$    \\
  LPIPS$\downarrow$          &       $0.152^{+0.017}_{-0.024}$  &       $0.152^{+0.017}_{-0.024}$ &\cellc$0.150^{+0.020}_{-0.024}$&    $0.199^{+0.027}_{-0.025}$ &       $0.210^{+0.025}_{-0.023}$ \\
  \midrule
  Param. [M]             &        $0.74$                    &       $0.21$                    &  $0.21$                    &       -                         &       -                         \\
  Runtime [ms]               &        $89.5\pm1.7$                     &       $30.9\pm1.2$                      &  -                         &       -                         &       -                         \\
  GMACs               &        $672$                     &       $175$                      &  $175$                         &       -                         &       -                         \\
  \bottomrule
\end{tabular}
\end{table}

The EASF and FSR1 upscalers achieve the highest average scores on common metrics like the PSNR and SSIM. This aligns with our previous observations that these models excel at reconstructing single-value summary statistics, which is important to capture the high-frequency noise grain or smoothness of the ground truth image. Hence, accurate reconstruction of these structural properties increases the SSIM and on average brings down the mean squared error (MSE) increasing the PSNR. However, as reflected by the 68\% confidence intervals in Table \ref{tab:metrics}, the differences in PSNR and SSIM scores might not be statistically significant compared to the LCS models.

Moreover, reconstructing simple summary statistics of the image is insufficient if we want to produce perceptually pleasing images. This is evident in the NIQE and LPIPS scores, where the LCS and its derived versions outperform both EASF and FSR1. The LCS models produce images that are perceptually satisfying at the expense of the PSNR and SSIM metrics. The NIQE and LPIPS scores appear to show the only statistically significant differences as evidenced by the non-overlapping confidence intervals in Table \ref{tab:metrics}.

In summary, we find that maximizing the PSNR and SSIM metrics does not necessarily lead to perceptually satisfying upscaled images \citep{Pathak_2016_blurryloss,Isola_2017_blurryloss}. As is also demonstrated in Figure~\ref{fig:appendix_comparison} of Appendix~\ref{appendix:figs}, which compares the HR ground truth, LR upscaled with a bicubic filter, LCS, and LCS trained with L1 loss only. 
Meanwhile, NIQE and LPIPS metrics are more aligned with our subjective evaluation of the upscaled images. Meanwhile, the JOD scores shows little variance across the board, making it a less useful metric for our purposes.

\section{Discussion and conclusions} \label{sec:disc_conc}
In this work we proposed a low-complexity scaler (LCS) trained on the GameIR dataset \citep{Zhou_2025_gameir}. We applied adversarial training to encourage reconstruction of perceptually important details, and used reparameterization and quantization techniques to reduce model complexity and size. Our comparative analysis with the EASF and FSR1 upscaling algorithms shows that our LCS achieves better performance in SISR on perceptual metrics such as NIQE and LPIPS, demonstrating the potential of AI-based ESR models for upscaling. With the exception of NIQE and LPIPS, the image quality metrics tested struggle to find statistically significant differences between the LCS, EASF and FSR1. Thus, we emphasize the importance of identifying suitable metrics, and encourage future extensions to explore a wider range of image quality metrics to identify potential objectives that will produce objectively better models. In addition, a real-time content upscaler like the LCS should eventually be evaluated on the perceptual quality and temporal stability of reconstructed moving images (i.e., video), as this would ultimately be the target use case for such a model.

Another direction for future work could focus on curating a more diverse training and validation dataset. As this would be a significant undertaking, another option would be to explore the possibility of unsupervised or self-supervised training on a larger, but potentially still limited, dataset of LR-HR image pairs. A larger model that adjusts parameters based on content metadata (e.g., game engine, rendering settings) could also merit further investigation. 

Finally, the efficiency and inference speed of the LCS could be accelerated on an NPU or another low-power device. Such an implementation could help free up GPU and system resources, thus improving game performance.
\clearpage

\begingroup
\let\itshape\upshape
\bibliographystyle{abbrv}
\bibliography{main.bib}
\endgroup

\appendices
\section{Appendix A: Additional Figures}\label{appendix:figs}
\begin{figure*}
  \centering
  \includegraphics[width=0.8\textwidth]{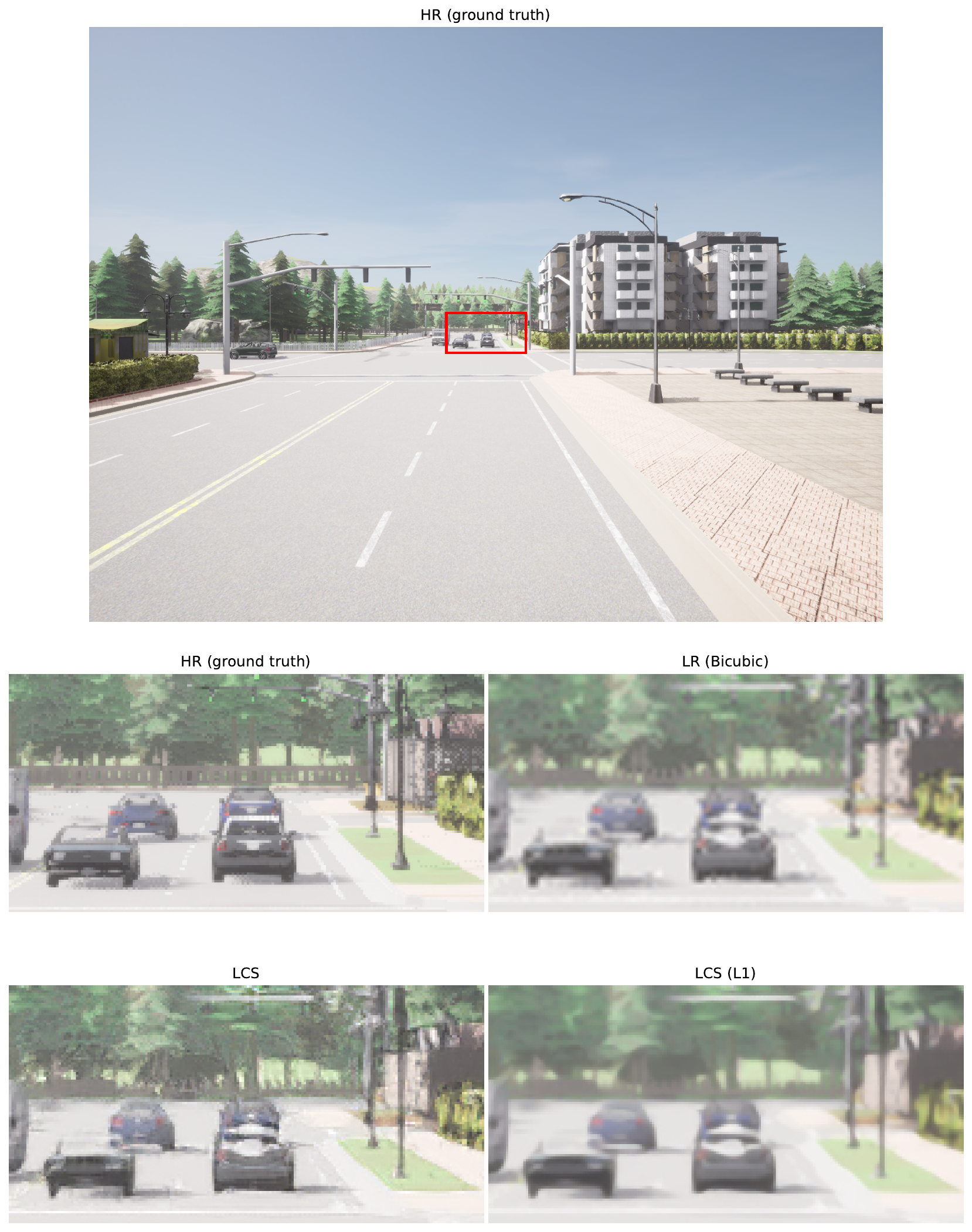}
  \caption{Comparison of the HR ground truth, LR upscaled with a bicubic filter, LCS, and LCS trained with L1 loss only. The LCS trained with L1 loss noticeably produces blurry images compared to the adversarially trained LCS. 
  }\label{fig:appendix_comparison}
\end{figure*}

\end{document}